\documentclass[conference]{IEEEtran}
\IEEEoverridecommandlockouts
\usepackage{cite}
\usepackage{amsmath,amssymb,amsfonts}
\usepackage{graphicx}
\usepackage{booktabs} 
\usepackage{longtable}
\usepackage{supertabular}
\usepackage{easyReview}
\usepackage{listings}
\usepackage{textcomp}
\usepackage{xcolor}
\usepackage{pythonhighlight}
\usepackage{multicol}
\usepackage{tabularray}  
\usepackage{amssymb}
\usepackage{lipsum}
\usepackage[ruled,vlined]{algorithm2e}
\usepackage[noend]{algpseudocode}
\usepackage{amsmath}
\usepackage{multirow}
\usepackage{easyReview}
\usepackage{soul}
\usepackage{arydshln}
\makeatletter
\newcommand{\linebreakand}{%
  \end{@IEEEauthorhalign}
  \hfill\mbox{}\par
  \mbox{}\hfill\begin{@IEEEauthorhalign}
}
\makeatother
\def\BibTeX{{\rm B\kern-.05em{\sc i\kern-.025em b}\kern-.08em
T\kern-.1667em\lower.7ex\hbox{E}\kern-.125emX}}
\usepackage{todonotes}
\usepackage{draftwatermark}
\SetWatermarkText{Accepted in IEEE IRI 2025}
\SetWatermarkScale{0.4}        % Adjust size
\SetWatermarkColor[gray]{0.8} % Light gray color

\begin{document}
\title{
%-Prompt Ensemble with LLMs for Reliable Medical Entity Recognition from EHRs

LLM-based Prompt Ensemble for Reliable Medical Entity Recognition from EHRs

% \thanks{Funding Info*}
}
\IEEEoverridecommandlockouts
\IEEEpubid{\begin{minipage}[t]{\textwidth}\ \\[10pt]
%\centering\normalsize{copyright-placeholder}
\end{minipage}}

\author{\IEEEauthorblockN{K M Sajjadul Islam}
\IEEEauthorblockA{\textit{Marquette University} \\
% \textit{Marquette University}\\
sajjad.islam@marquette.edu}
\and
\IEEEauthorblockN{Ayesha Siddika Nipu}
\IEEEauthorblockA{\textit{University of Wisconsin-Milwaukee} \\
nipua@uwm.edu}

\linebreakand
\IEEEauthorblockN{Jiawei Wu}
\IEEEauthorblockA{\textit{Medical College of Wisconsin} \\
jiawu@mcw.edu}
\and
\IEEEauthorblockN{Praveen Madiraju}
\IEEEauthorblockA{\textit{Marquette University} \\
% \textit{Marquette University}\\
praveen.madiraju@marquette.edu}
}
\maketitle 

\begin{abstract}
Electronic Health Records (EHRs) are digital records of patient information, often containing unstructured clinical text. Named Entity Recognition (NER) is essential in EHRs for extracting key medical entities like problems, tests, and treatments to support downstream clinical applications. This paper explores prompt-based medical entity recognition using large language models (LLMs), specifically GPT-4o and DeepSeek-R1, guided by various prompt engineering techniques, including zero-shot, few-shot, and an ensemble approach. Among all strategies, GPT-4o with prompt ensemble achieved the highest classification performance with an F1-score of 0.95 and recall of 0.98, outperforming DeepSeek-R1 on the task. The ensemble method improved reliability by aggregating outputs through embedding-based similarity and majority voting.
\end{abstract}


\section{Introduction}
Electronic Health Records (EHRs) are digital systems that store and manage patient information across clinical encounters. Initially designed to support administrative tasks like billing and scheduling, EHRs have now become essential tools in healthcare delivery and research. Over the past two decades, adoption of EHRs has grown rapidly, especially in the United States \cite{kim2019evolving}. Beyond routine clinical use, EHRs enable secondary applications such as real-world evidence generation, disease surveillance, and clinical trial design. However, the unstructured and fragmented nature of much of the data requires advanced techniques such as natural language processing (NLP) and machine learning to make it usable for research.

\begin{figure}[tb]
\centerline{\includegraphics[width=0.35\textwidth]{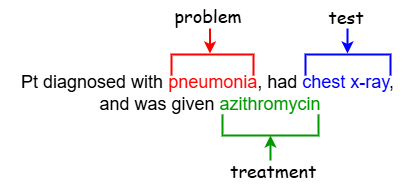}}
\caption{Example of entities in EHR}
\label{Fig:sample_medical_record}
\end{figure}

Named Entity Recognition (NER) is a fundamental task in NLP that involves identifying and categorizing key entities in text into predefined categories such as problems, tests, and treatments as illustrated in Figure~\ref{Fig:sample_medical_record}. In the context of EHR, NER plays a crucial role by extracting clinically relevant information from unstructured text such as physician notes and discharge summaries. This structured information enables more efficient indexing, retrieval, and analysis of patient data. Accurate medical entity recognition supports downstream applications like clinical decision support, cohort identification, and disease surveillance, thereby improving patient care and healthcare research outcomes\cite{yadav2018mining}.

Despite its importance, clinical NER presents many challenges due to the nature of medical language and documentation. Clinical notes are often unstructured, written in narrative form, and filled with domain-specific terminology, abbreviations, and inconsistencies. The same abbreviation may have multiple meanings depending on context, and many entity names overlap or have ambiguous boundaries. Variability in writing styles and the lack of standard document structure across institutions further complicate automated extraction. Additionally, accurate entity recognition requires handling negation, disjoint mentions, and classification of findings into actionable or historical categories \cite{kundeti2016clinical}. These issues make NER in clinical text a non-trivial task and necessitate robust, adaptable methods.

Large Language Models (LLMs) have shown strong potential in performing NER through prompt-based learning. Unlike traditional models that require fine-tuning, LLMs can identify entities using natural language instructions and examples. Recent studies have demonstrated that LLMs can achieve competitive results on general NER tasks with minimal supervision \cite{wang2023gpt}. In the clinical domain, LLMs have also been successfully applied to extract medical entities from EHRs. When guided by well-designed prompts, LLMs like GPT-4 can approach the performance of specialized models trained on medical data, offering a scalable and flexible solution for clinical information extraction \cite{hu2024improving}. In this work, we pursue two main research aims:

\textbf{Aim 1} is to investigate the effectiveness of prompting strategies, including zero shot, few shot, and ensemble for medical entity recognition using GPT-4o and DeepSeek-R1.

\textbf{Aim 2} is to systematically evaluate the performance of GPT-4o and DeepSeek-R1 on clinical entity recognition tasks by measuring entity extraction and classification performance across different prompting strategies.

\section{Related Work}
\subsection{Traditional Approaches for Medical NER}\label{sec-traditional-models}
Early approaches to recognition of medical entities relied heavily on rule-based systems and expert-crafted lexicons to identify clinical concepts such as medications, symptoms and diagnoses. These systems demonstrated good precision in restricted domains, but were time-consuming to maintain and lacked adaptability to unseen data \cite{bose2021survey}. To overcome these limitations, statistical machine learning models like Conditional Random Fields (CRFs) and Support Vector Machines (SVMs) were introduced, treating NER as a sequence labeling task \cite{fu2020clinical}. These models required manual feature engineering and were limited in handling long-range dependencies in text.

With the advancement of deep learning, neural models such as BiLSTM-CRF began to dominate clinical NER tasks by automatically learning contextual features from annotated corpora \cite{bose2021survey}. More recently, transformer-based models have significantly improved performance in medical NER. Pretrained language models like BERT and its biomedical variants such as BioBERT \cite{yu2019biobert}, ClinicalBERT \cite{fu2020clinical}, and Med-BERT \cite{liu2021med}, leverage large domain-specific corpora and deep contextual encoding to achieve results on clinical entity recognition benchmarks.

\subsection{LLMs for Downstream Clinical Applications}\label{sec-clinical-task}
LLMs, such as GPT-4 and Claude, have emerged as powerful tools capable of understanding and generating human-like text. These models are increasingly applied in the medical domain because they can process unstructured clinical data and support decision-making through prompt-based interactions. Prompting allows users to communicate with LLMs in natural language, enabling flexible and context-aware outputs without requiring task-specific fine-tuning. Studies show that LLMs can interpret patient complaints, summarize clinical records, and assist with disease classification through carefully designed prompts \cite{schmiedmayer2024llm}. For instance, GPT-4 has demonstrated strong performance in classifying patient complaints using few-shot prompts, improving accuracy as more training examples are provided \cite{nipu2024reliable}. Moreover, applications like LLM on FHIR leverage prompting to translate complex medical records into patient-friendly language, improving health literacy and access to care\cite{schmiedmayer2024llm}. Despite their promise, the variability in LLM responses and risks of hallucination underline the importance of rigorous evaluation and context filtering in clinical use.

\subsection{LLM-based Prompting Approaches for Clinical NER}\label{sec-llm-prompting}
Prompting has emerged as a flexible alternative to fine-tuning in clinical NER, reducing the need for large annotated datasets. Recent studies show that language models like GPT-3.5 and GPT-4 can perform clinical NER when guided by well-crafted prompts. Hu et al. proposed a structured prompting framework that combines task instructions, annotation guidelines, and few-shot examples. Their results showed that GPT-4, when properly prompted, achieved competitive performance with domain-specific models like BioClinicalBERT on datasets such as MTSamples and VAERS \cite{hu2024improving}. Chen et al. introduced KGPC, a method that frames NER as a question answering task using knowledge-guided prompts and contrastive learning. Their model outperformed existing few-shot baselines by leveraging UMLS-based entity generation and prompt tuning \cite{chen2023few}. Shen et al. evaluated ChatGPT for extracting structured data from clinical notes using few-shot prompting and evidence-based reasoning. Their approach surpassed CRF and BERT-based NER models, highlighting the potential of prompt-based inference for real-world clinical applications \cite{huang2024critical}. Together, these studies demonstrate that prompt engineering enables general-purpose language models to perform robust clinical NER without task-specific training.

Many recent studies in prompt-based medical NER rely on benchmark datasets such as the 2010 i2b2/VA Natural Language Processing Challenge\cite{uzuner20112010}, which provides standardized clinical annotations for evaluating entity extraction and classification. While prompting strategies have shown promise, most works focus on individual prompt designs or specific evaluation metrics. To the best of our knowledge, no prior study has explored a prompt ensemble approach with a robust evaluation across both entity extraction and classification. Our work is among the first to apply such an approach to state-of-the-art LLMs, including GPT-4o and DeepSeek-R1, highlighting the importance of prompt reliability and consistency in clinical entity recognition.

\begin{table}[tb]
\caption{Frequent Clinical Entities in Dataset}
  \label{table-entities}
  \centering
  \begin{tabular}{lll}
    \toprule
     Entity Type & Most Frequent & Least Frequent \\
    \midrule
    Problem & hypertension & minimal ooze \\
    Test &  glucose & CIWA scale \\
    Treatment & aspirin & pneumatic boots\\
    \bottomrule
  \end{tabular}
\end{table}

\section{Methodology}
\subsection{Dataset Description}\label{sec-dataset}
We used the publicly available dataset from the 2010 i2b2/VA Natural Language Processing Challenge\cite{uzuner20112010}, which focuses on extracting concepts, assertions, and relations from clinical text. The dataset comprises de-identified EHRs, primarily discharge summaries, annotated with three types of clinical entities: Problem, Test, and Treatment. Each entity is marked with its span in the text and categorized according to its clinical role. Problems represent diseases or symptoms, Tests indicate diagnostic procedures, and Treatments refer to medications or interventions. Table~\ref{table-entities} presents examples of the most and least frequent entities for each category.

To support prompt-based medical entity recognition, we constructed a compact dataset tailored to each prompting strategy, as extensive data is not required for LLMs. We employed five prompting techniques to extract and classify entities (more details on section \ref{sec-Prompt-Engineering}). For training, we sampled one full document, 100 labeled sentences from five documents, and all available entities from 73 documents, depending on the prompting technique. We used a separate document with 190 annotated entities as the test sample. Table~\ref{table-datasets} summarizes the distribution of entity types used across different prompting configurations.

\begin{table}[tb]
\caption{Entity Distribution for Different Prompting}
  \label{table-datasets}
  \centering
  \begin{tabular}{lllll}
    \toprule
     Prompt & Samples & Problem & Test & Treatment\\
    \midrule
    Zero-shot & No sample & - &  - & -\\
    Few-shot 1 & Single Doc & 53 & 54 & 52 \\
    Few-shot 2 & 100 Sentences$^{\mathrm{a}}$ & 60 & 90 & 71 \\
    Few-shot 3 & All Entities$^{\mathrm{b}}$ & 2567 &  1206 & 1582 \\
    \hline
    Test Sample & Single Doc & 72 & 66 & 52 \\
    \bottomrule
  \end{tabular}
\flushleft
  {\footnotesize $^{\mathrm{a}}$Collected from 5 documents, $^{\mathrm{b}}$Collected from 73 documents}
\end{table}

\subsection{Prompt Engineering}\label{sec-Prompt-Engineering}
We utilized two LLMs, GPT-4o \cite{hurst2024gpt} and DeepSeek-R1 \cite{guo2025deepseek}, to perform medical entity recognition directly through prompt-based interaction. While these models can be accessed via user interfaces or APIs, in this study, we integrated them using API endpoints to enable automated querying and result collection across various prompting strategies. We configured the generation parameters with a $temperature$ of 0.2 and a $top\_p$ value of 1 to promote deterministic and focused outputs while still allowing a limited degree of sampling diversity, ensuring consistency across multiple runs without completely eliminating variation.

Prompt engineering is a key methodological component in this work, enabling us to harness the reasoning capabilities of LLMs without any fine-tuning\cite{alammar2024hands}. It involves crafting input prompts that clearly communicate the task, including structured instructions, relevant context, and expected output formats. Effective prompts help LLMs understand both what to do and how to do it, guiding them to extract and categorize entities from clinical text. In this study, we experimented with various prompting strategies that differ in the amount and type of supervision provided, such as instructions alone or examples at different levels of granularity. These strategies are detailed in the following subsections.

\begin{figure}[tb]
\centerline{\includegraphics[width=0.40\textwidth]{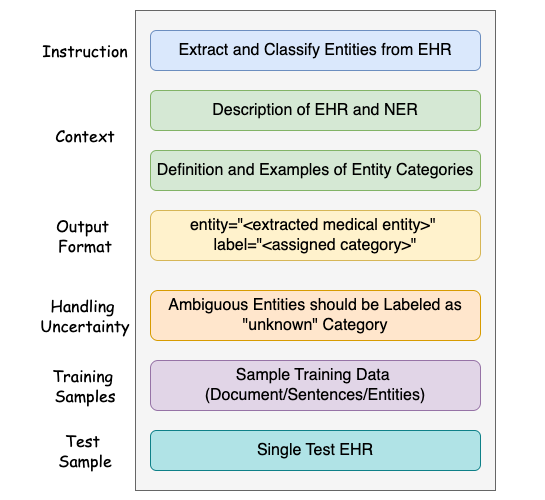}}
\caption{Components of Prompting Template}
\label{Fig:Prompting-Template}
\end{figure}

To ensure consistent and interpretable interactions with the language models, we designed a structured prompt template. It includes clear instructions, relevant context, and formatting cues, as shown in Figure~\ref{Fig:Prompting-Template}. The prompt begins with a task definition that focuses on both entity extraction and classification. This is followed by context describing the clinical data source, specifically EHRs used in NER. The prompt then introduces the three target categories: Problem, Test, and Treatment. Each category is explained with examples to help the model understand their roles in clinical text. We also define the expected output format to reduce ambiguity in the model’s response. This format also makes it easier to evaluate the model’s performance. In addition, we instruct the model to assign an ``unknown" label when it cannot confidently classify an entity. This helps reduce hallucinations or incorrect assumptions.
We designed the template to be concise and modular to address the issue of models forgetting information in the middle of long prompts \cite{liu2024lost}. 
For few-shot prompting, we included training samples using three different formats: document, sentence, and entity level. This allowed us to examine how different levels of context affect the model’s ability to extract and classify medical entities.
During the prompt template development, we performed an `ablation study' \cite{wei2022chain} comparing various prompt structures and sample configurations. Based on the evaluation, we selected the most effective template and input format for our prompting strategies.

\begin{algorithm}[tb]
    \SetAlgoLined
    \caption{Prompt Ensemble with Majority Voting}
    \label{algMajorityVoting}
    \textbf{Input:} \\
    $\mathcal{P} = \{p_1, p_2, p_3, p_4\}$ \textcolor{gray}{// entity-label pair from 4 prompts} \\
    $\tau=0.92$ \hspace{0.2cm} \textcolor{gray}{// cosine similarity threshold}
    
    \textbf{Output:} Entity predictions with majority-voted labels
    
     \textcolor{gray}{// entity embedding} 
    
    \For{each entity in $p_i \in \mathcal{P}$}{
        Compute ClinicalBERT embedding for \texttt{entity}
        
        Add \{entity, label, embedding\} to $\mathcal{E}$
    }
     \textcolor{gray}{// greedy hierarchical clustering}
    
    \For{each entity $e_i \in \mathcal{E}$}{
        Compute cosine similarity $s$ between entities \\
        \If{$s \geq \tau$}{
            Add to cluster $\mathcal{C}$
        }
    }
    
    \textcolor{gray}{// majority vote count}
    
    \For{each cluster $c \in \mathcal{C}$}{
        Count label frequencies $f$ in $c$ \\
        \eIf{$f >= 2$ }{
            Assign majority label
        }{
            Assign label as \texttt{unknown}
        }
        Add \texttt{entity, label} to output $\mathcal{O}$
    }
    %\Return{$\mathcal{O}$}
\end{algorithm}

\textbf{Zero-shot prompting} involves providing the model with task instructions and label definitions without any annotated examples, relying entirely on its pretrained knowledge to perform the task\cite{islam2023autocompletion}. In this setting, the prompt includes a concise task description, background context, definitions for the entity categories (Problem, Test, Treatment), and formatting expectations. The clinical text is then appended at the end of the prompt. This approach serves as a baseline to evaluate the model’s ability to understand and execute the task without seeing any in-context examples.

\textbf{Few-shot prompting with a document sample} involves providing the model with a single annotated clinical document to help it learn how to extract and categorize entities from context, where medical entities are tagged using XML-style tags such as $<problem>$, $<test>$, and $<treatment>$. Additional instructions clarify that not all sentences may contain entities, and some may include multiple entities from different categories. This training sample is enclosed within triple quotes and inserted before the test input. The goal of this setup is to allow the model to learn entity patterns from a realistic, document-level context and apply similar reasoning to unseen clinical text.

\textbf{Few-shot with sentence samples} uses a set of 100 annotated clinical sentences collected from 5 documents as training input. Medical entities in each sentence are labeled using XML style tags to indicate their respective categories. Unlike the document level approach, this format excludes unrelated or unlabeled sentences, helping the model focus only on relevant patterns. The goal is to provide clear, sentence level guidance to improve the model’s ability to generalize entity recognition logic when processing new clinical text.

\textbf{Few-shot with entity samples} provides the model with a curated list of labeled medical entities, grouped by their respective categories. These entities are presented in a flat, comma-separated format enclosed within triple quotes, without full sentence or document context. Each entity is tagged using XML style tags to indicate its classification. We supplied a large number of training entities, 5,355 in total, collected from 73 clinical documents. This included 2,567 Problem, 1,206 Test, and 1,582 Treatment entities. Although this approach is straightforward, it is significantly longer compared to previous approaches due to the size of the entity list. The document and sentence level few-shot strategies teach the model how entities appear within context. This method focuses on showing the model individual medical entities without any surrounding text, emphasizing category-specific terminology.

\begin{figure}[tb]
\centerline{\includegraphics[width=0.416\textwidth]{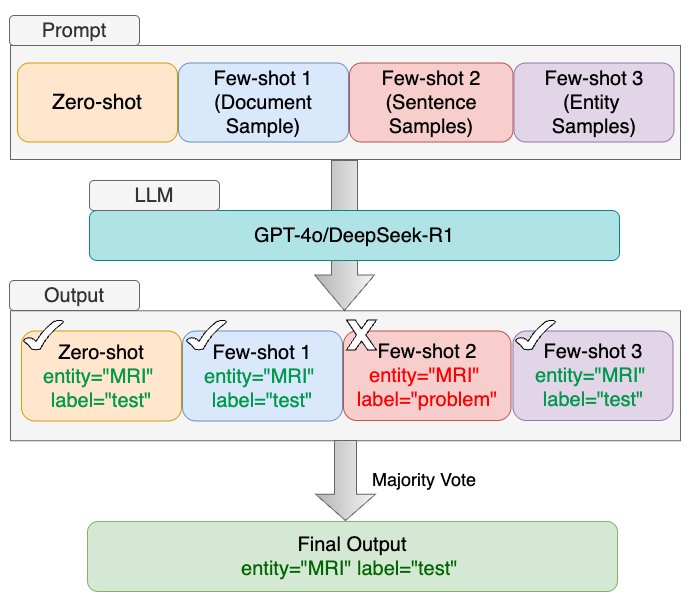}}
\caption{Workflow of the Prompt Ensemble}
\label{Fig:prompt-ensemble}
\end{figure}

\textbf{Prompt ensemble} is employed as our final strategy to improve prediction consistency by aggregating outputs from multiple prompt formats. We run three different few-shot configurations using different types of training samples: document, sentences, and entities, each offering a distinct view of the data. After extracting entities from each prompt, we embed them using ClinicalBERT\cite{alsentzer2019publicly} and match similar entities utilizing cosine similarity. For each semantically similar entity appearing in at least two prompt outputs, we assign a final label using majority voting; otherwise, the label is set to ``unknown''. The detailed steps of this ensemble prompting strategy are outlined in Algorithm~\ref{algMajorityVoting}. Examples of matched entity pairs with their cosine similarity scores are shown in Table~\ref{table-entity-sim-matching}, illustrating the effectiveness of embedding-based entity alignment. This embedding-based ensemble approach mitigates prompt level variation and enhances robustness in medical entity recognition. An overview of this ensemble prompting workflow is illustrated in Figure~\ref{Fig:prompt-ensemble}.

\begin{table}[tb]
\caption{Entity Similarity Matching Samples}
  \label{table-entity-sim-matching}
  \centering
  \begin{tabular}{lll}
    \toprule
     Actual Entity & Extracted Entity & Similarity Score \\
    \midrule
    lower abdominal pain & lower abd pain & 0.98 \\
    angiography & angiogram & 0.96 \\
    UREA N & urea n-25 & 0.94 \\
    \hdashline
    colitis & ulcer & 0.91 \\
    fluid & URINE & 0.90 \\
    \bottomrule
  \end{tabular}
\end{table}

\section{Result Analysis And Discussion}

This section presents a comprehensive evaluation of our experimental findings. We analyze performance across several dimensions: findings from similarity measurement, entity extraction and classification performance, computational efficiency, and ethical considerations related to prompt design. Our discussion not only interprets the metrics, but also reflects on model behavior and the practical implications of design choices in prompt-based medical entity recognition.

We used ClinicalBERT embeddings with cosine similarity to identify \textbf{entity similarity} extracted by different prompting strategies. We applied a similarity threshold of 0.92, which was empirically chosen to keep the matches accurate while still covering enough relevant entities. Unlike simple string matching, which fails to capture variations in clinical terminology and abbreviations, it often misses semantically related terms. Embedding-based similarity considers the meaning of the text, enabling more robust matching of entities with different surface forms. As shown in Table~\ref{table-entity-sim-matching}, pairs such as ``lower abdominal pain" and ``lower abd pain" (similarity: 0.98), and ``angiography" and ``angiogram" (0.96) are correctly matched. Conversely, less precise pairs like ``fluid" and ``urine" (0.90) fall below the threshold. This design decision ensures only well-aligned predictions are grouped during ensemble voting, helping reduce hallucinations and label noise in final outputs. It also improves the quality of entity classification evaluation by ensuring that only semantically consistent entity mentions are assigned a final label.

\begin{table}[tb]
\caption{Entity Extraction Performance}
  \label{table-entity-extraction}
  \centering
  \begin{tabular}{llllll}
    \toprule
     Model & Prompt & Predict & Match & Unknown & Accuracy \\
    \midrule
    \multirow{5}{*}{GPT-4o} & Zero-shot  & 76 &  71 & 0 & 0.37 \\
    & Few-shot 1  & 121 & 108 & 7 & 0.56 \\
    & Few-shot 2  & 117 & 112 & 0 & 0.59 \\
    & Few-shot 3 & 139 &  123 & 0 & \textbf{0.65} \\
    & Ensemble  & 75 &  70 & 0 & 0.37 \\
     \hline
    \multirow{5}{*}{DeepSeek-R1} & Zero-shot  & 54 & 50 & 3 & 0.26 \\
    & Few-shot 1  & 84 & 79 & 0 & 0.41 \\
    & Few-shot 2  & 106 & 99 & 0 & 0.52 \\
    & Few-shot 3 & 58 &  53 & 0 & 0.28 \\
    & Ensemble  & 49 & 42 & 3 & 0.22 \\
    \bottomrule
  \end{tabular}
\end{table}

Table~\ref{table-entity-extraction} presents the \textbf{entity extraction performance} of GPT-4o and DeepSeek-R1. For GPT-4o, we observe performance differences across prompting formats, with the highest accuracy of 65\% achieved using Few-shot 3. This setup includes a comprehensive list of training entities (excluding those present in the test document), which likely helps the model recognize similar entities through exposure to extensive medical terminology. Zero-shot and ensemble prompting both resulted in an accuracy of 37\%. While this lower performance is expected in the absence of training examples for zero-shot, ensemble prompting’s result reflects the conservative nature of majority voting. Although less accurate, the ensemble output is more reliable due to its stricter agreement criteria. In contrast, DeepSeek-R1 showed generally lower performance. Its best result, 52\% accuracy, was achieved using Few-shot 2, where the prompt includes a set of annotated clinical sentences. Ensemble prompting further reduced performance for DeepSeek-R1, highlighting its inconsistency across different prompting strategies. Overall, GPT-4o consistently outperformed DeepSeek-R1, demonstrating stronger generalization capabilities in prompt-based clinical entity extraction.

\begin{table}[tb]
\caption{Entity Classification Performance}
  \label{table-entity-classification}
  \centering
  \begin{tabular}{llllll}
    \toprule
     Model & Prompt & Precision & Recall & F1 \\
    \midrule
    \multirow{5}{*}{GPT-4o} & Zero-shot  &  0.88 & 0.94 & 0.91 \\
    & Few-shot 1  & 0.92 & 0.97 & 0.94 \\
    & Few-shot 2  & 0.92 & 0.96 & 0.94 \\
    & Few-shot 3  &  0.81 & 0.92 & 0.86 \\
    & Ensemble    & 0.92 & \textbf{0.98} & \textbf{0.95} \\
     \hline
    \multirow{5}{*}{DeepSeek-R1}& Zero-shot &  0.90 & 0.92 & 0.91 \\
    & Few-shot 1  & 0.92 & 0.97 & 0.94 \\
    & Few-shot 2  & 0.89 & 0.96 & 0.93 \\
    & Few-shot 3  &  0.84 & 0.92 & 0.88 \\
    & Ensemble    & 0.85 & 0.93 & 0.89 \\
    \bottomrule
  \end{tabular}
\end{table}

Table~\ref{table-entity-classification} summarizes \textbf{entity classification performance} of GPT-4o and DeepSeek-R1 across various prompting strategies, based on precision, recall, and F1 score. For GPT-4o, the zero-shot configuration achieves a strong F1 score of $0.91$, which further improves with Few-shot 1 approach when a clinical document was given as the training sample. 
The ensemble prompting strategy for GPT-4o achieves the highest F1 score of 0.95, highlighting its effectiveness in capturing a broader set of true positive entities.
Although entity extraction performance is low, the classification F1-score surpasses the 0.924 achieved by the i2b2/VA concept extraction system \cite{uzuner20112010}.
The recall achieved by GPT-4o is 0.98 for ensemble prompting, indicating strong sensitivity in identifying relevant clinical entities. Recall, also known as sensitivity, reflects the model's ability to correctly identify actual positive cases. A higher recall helps minimize the chance of overlooking important clinical information, which is especially critical in healthcare applications.
%The ensemble prompting strategy for GPT-4o achieves the highest recall of 0.98 and an F1 score of 0.95, highlighting its effectiveness in capturing a broader set of true positive entities. 
%This F1 score surpasses the best-performing concept extraction system from the 2010 i2b2/VA NLP Challenge, which achieved an exact F1-score of 0.852 and an inexact F1-score of 0.924 using a semi-supervised CRF-based approach \cite{uzuner20112010}.
%As recall, which is also known as sensitivity, measures the proportion of actual positives that are correctly identified, it indicates how effectively the model captures relevant entities. A high recall value reflects the model's ability to identify most clinical concepts, such as problems, tests or treatments, thereby reducing the risk of missing critical information that could negatively impact downstream decision-making. 
In contrast, DeepSeek-R1 demonstrates stable performance, with a zero-shot F1 score of 0.91 and its highest score of 0.94 achieved under Few-shot 1. However, its performance declines under Few-shot 3, yielding an F1 of 0.89 despite a relatively high recall of 0.93. These results suggest that while DeepSeek-R1 benefits from moderate few-shot supervision, it is more sensitive to prompt length and complexity. Overall, GPT-4o consistently achieves higher recall and F1 across strategies, making it more reliable for comprehensive clinical entity classification tasks.

\begin{figure}[tbp]
\centerline{\includegraphics[width=0.35\textwidth]{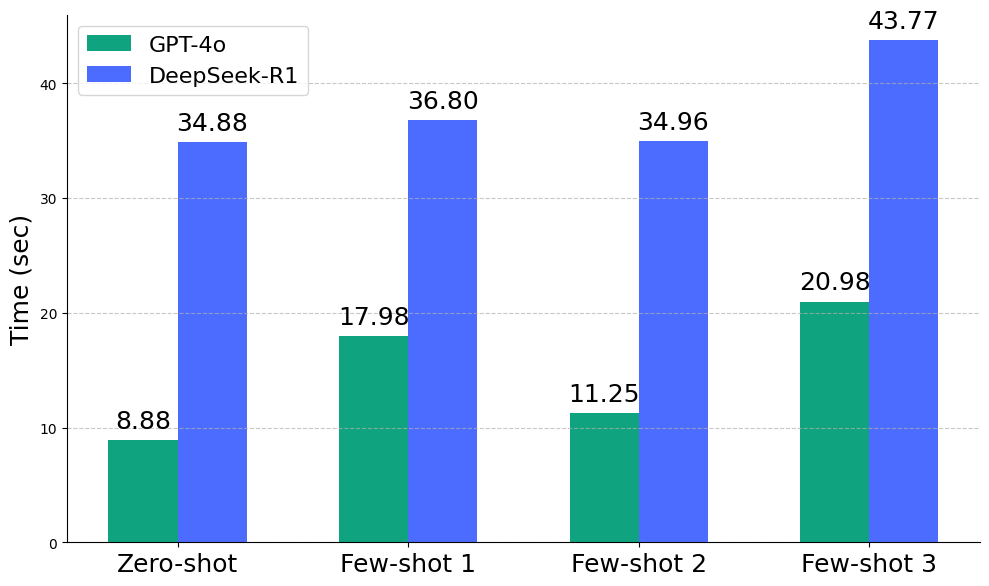}}
\caption{Prompt Execution Time}
\label{Fig:execution-time}
\end{figure}

We further analyze the \textbf{computational efficiency} and prompt length sensitivity of both models. As shown in Figure~\ref{Fig:execution-time}, GPT-4o consistently executes faster than DeepSeek-R1, with the most notable difference in the zero-shot setting, where GPT-4o completes the task in 8.88 seconds compared to 34.88 seconds for DeepSeek-R1. In the Few-shot 3 configuration, which includes all entity list as training sample, execution time increases significantly for both models. GPT-4o takes 20.98 seconds, while DeepSeek-R1 takes 43.77 seconds. This increase in execution time is primarily due to the larger number of tokens in the Few-shot 3 configuration compared to other prompting techniques. Longer prompts result in more tokens being processed, which leads to longer response generation time. GPT-4o, despite stronger performance, triggered a token limit error when the prompt exceeded the model’s tokens-per-minute (TPM) allowance. To address this, we reduced the input size by trimming 10\% of entities from each category. It ensures that the prompt remains within the model token limits while preserving balanced representation across entity types. These findings highlight the importance of prompt efficiency and model-specific tuning when designing few-shot configurations, particularly for large-scale or time-sensitive clinical applications.

\textbf{Ethical Consideration:}
In our study, we used the publicly available and anonymized dataset from the 2010 i2b2/VA NLP Challenge, which removes risks related to patient identification as the records are anonymized. However, when deploying LLMs via third-party APIs, one should consider possible data leakage as clinical data transmitted externally may be exposed without proper safeguards. Recent studies have also highlighted key ethical concerns about LLM hallucination, where models generate plausible but incorrect outputs. Huang et al. noted that ChatGPT occasionally fabricated clinical values when data was missing, instead of indicating uncertainty, which could compromise patient safety if used unsupervised \cite{huang2024critical}. To address this, we introduced an ``unknown" label in our prompting strategy, allowing the model to abstain from classification when confidence is low. While this helped reduce hallucination, it still required human judgment to review the outcome \cite{nipu2024reliable}.

\section{Conclusion}
This study explored prompt-based medical entity recognition from EHRs using two LLMs, GPT-4o and DeepSeek-R1. Among the evaluated prompting strategies, GPT-4o consistently outperformed DeepSeek-R1 across both entity extraction and classification tasks. The prompt ensemble method demonstrated improved reliability by aggregating predictions from multiple prompt formats and using embedding-based similarity with majority voting to reduce hallucination and label noise. This approach enhances the robustness and consistency of LLMs in clinical entity recognition, making them more dependable for extracting structured information from unstructured medical text.

\section*{Acknowledgment}
We gratefully acknowledge the support of the Northwestern Mutual Data Science Institute for funding this project.

\bibliographystyle{IEEEtran}
\bibliography{Ref}
\vspace{12pt}

\end{document}